\pdfoutput=1

\documentclass[11pt]{article}

\usepackage[]{EMNLP2023}

\usepackage{times}
\usepackage{latexsym}

\usepackage[T1]{fontenc}

\usepackage[utf8]{inputenc}

\usepackage{microtype}

\usepackage{inconsolata}

\usepackage{graphicx}
\usepackage{amsmath}
\usepackage{amssymb}
\usepackage{booktabs}

\usepackage{color, booktabs,subcaption,amsfonts,dcolumn}
\usepackage{amsfonts}
\usepackage{graphicx}

\usepackage{multirow,diagbox,makecell}
\usepackage{tikz}

\usepackage{pifont}
\newcommand{\cmark}{\ding{51}}%
\newcommand{\xmark}{\ding{55}}%
\usepackage{booktabs}

\usepackage{capt-of}

\title{\emph{R2H}: Building Multimodal Navigation Helpers that \\\emph{Respond to Help Requests}}

\author{Yue Fan, Jing Gu, Kaizhi Zheng, Xin Eric Wang \\University of California, Santa Cruz \\ \{yfan71, kzheng31, jgu110, xwang366\}@ucsc.edu}

\begin{document}
\maketitle
\begin{abstract}
Intelligent navigation-helper agents are critical as they can navigate users in unknown areas through environmental awareness and conversational ability, serving as potential accessibility tools for individuals with disabilities. In this work, we first introduce a novel benchmark, \emph{Respond to Help Requests (R2H)}, to promote the development of multi-modal navigation helpers capable of responding to requests for help, utilizing existing dialog-based embodied datasets. R2H mainly includes two tasks: (1) Respond to Dialog History (RDH), which assesses the helper agent's ability to generate informative responses based on a given dialog history, and (2) Respond during Interaction (RdI), which evaluates the effectiveness and efficiency of the response during consistent cooperation with a task performer. Furthermore, we explore two approaches to construct the navigation-helper agent, including fine-tuning a novel task-oriented multi-modal response generation model that can see and respond, named \emph{SeeRee}, and employing a multi-modal large language model in a zero-shot manner. Analysis of the task and method was conducted based on both automatic benchmarking and human evaluations. 
Project website: \url{https://sites.google.com/view/response2helprequests/home}.

\end{abstract}

\begin{figure}[t]
\centering
\setlength{\abovecaptionskip}{0.1cm}
\includegraphics[width = 7.5cm]{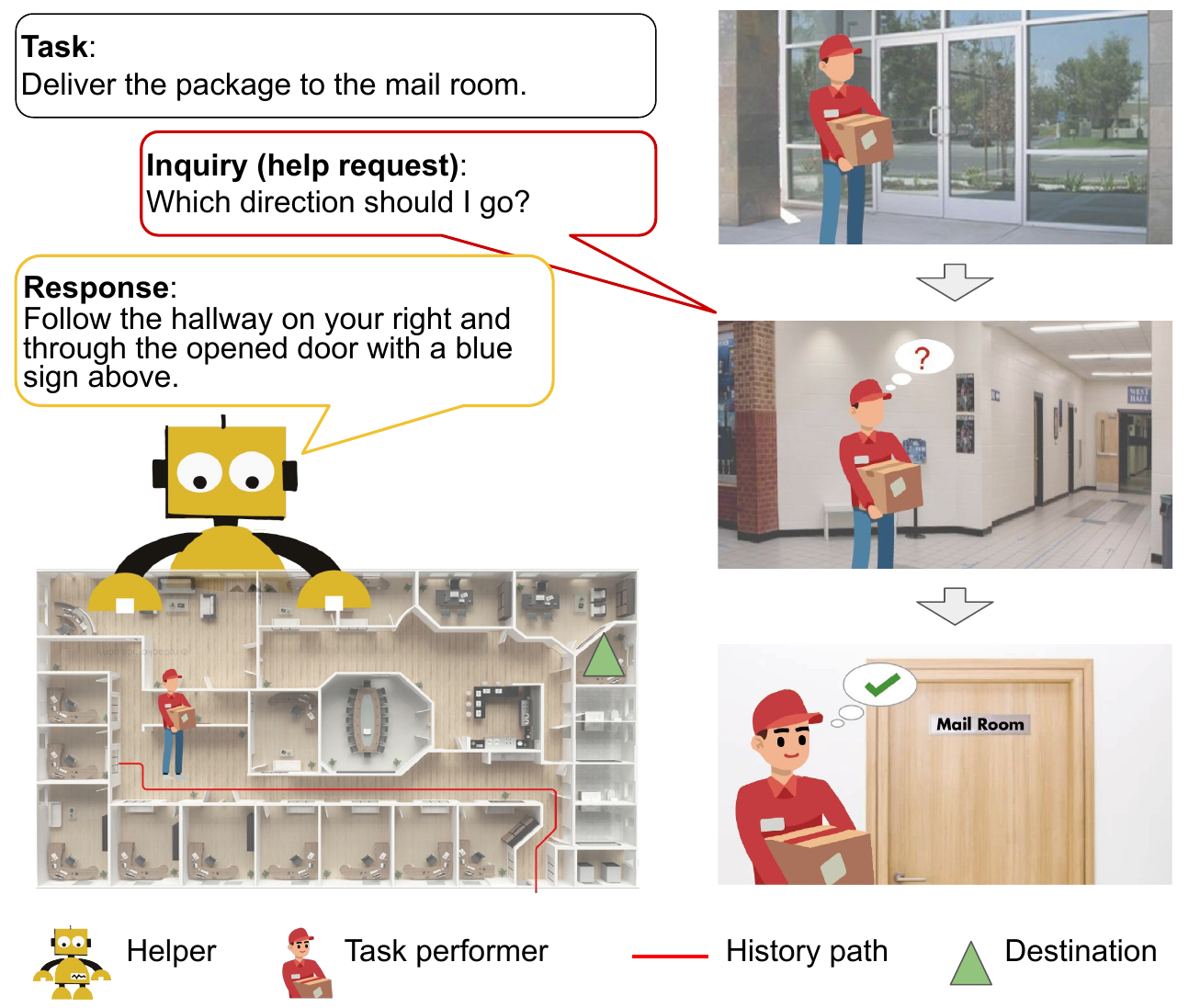}

\caption{Example of a helper agent. A navigation helper provides responses to help a task performer who is delivering a package. The helper has access to oracle information that is not available to the task performer, such as the location of the destination and map of the environment.
}
\label{figure1}
\end{figure}

\section{Introduction}

Assisting humans in real-world scenarios is a fundamental capability that AI agents should possess. An AI helper, communicating with humans in natural language based on visual observation in the environment and oracle information could significantly enhance work productivity and serve as an accessibility tool for individuals with disabilities. Figure~\ref{figure1} illustrates an example of a delivery person seeking assistance in navigating through an unfamiliar building. With the help of a navigation-helper agent, the delivery person can ask questions about directions and receive responses that are tailored to visual information about the current surroundings.

Building such a helper agent poses significant challenges. It requires understanding the visual environment and the task performer's inquiries and leveraging oracle information to provide effective responses. Evaluating these agents also requires a task performer to show how well the helper agent performs in the real collaborative real scenario, where the task performer follows instructions and further sends inquiries when needed to the helper with the goal of completing tasks. Using a human as an oracle task performer would be the most intuitive setting, but it is impractical due to the high cost and low efficiency. 

In this work, we introduce the Respond to Help Requests (R2H) benchmark, designed to automatically evaluate conversational multi-modal navigation helpers in a cooperative dynamic with another agent as the task performer. The R2H benchmark incorporates pre-trained performer agents to follow the responses from the helper agent, and the helper agent's performance is then reflected in the performance of the fixed task performer.
Leveraging three existing vision-and-dialog navigation datasets, CVDN \cite{thomason2020vision}, AlFRED \cite{ALFRED20} and AVDN \cite{fan2022aerial}, our R2H benchmark introduces two novel tasks: the Respond to Dialog History task (RDH) and the Respond during Interaction task (RdI). In the RDH task, the helper agent generates a response to the inquiry in the dialog history from humans, aiming at facilitating the task completion of the performer agent. In the RdI task, the helper agent needs to generate multiple responses from the start of the navigation process with no prior dialog history till the task's success.
To this end, R2H benchmark offers a pragmatic evaluation of the response from helper agents in both single- and multi-turn helper-performer cooperation.

We also present a multi-modal helper agent SeeRee for R2H benchmark, which leverages the oracle knowledge about the task and environment such as the destination location in a navigation task, to generate responses to inquiries from the task performer. 
SeeRee employs pre-trained vision and language models
to handle multi-modal inputs. To manage long input sequences, SeeRee leverages a novel Conditional Optimized Sparse (COS) attention mask. Moreover, we introduce a Parse by Step, which leverages Large Language Model to transform ground-truth human responses into structured step-by-step navigation instructions. Those parsed instructions (instead of human responses) serve as a better training source with improved performance in helping the task performer. In experiments, SeeRee surpasses the baseline in generating effective responses and validates the COS attention mask and Parse by Step method. SeeRee's responses have been evaluated through human assessments, demonstrating high accuracy and a significant improvement in task success rate compared to the baseline.



Additionally, we ask human testers to rate the faithfulness and naturalness of the responses evaluate the response from helper agents with automatic scores. As a result, our experiments indicate that a higher language similarity to human helpers does not necessarily lead to a more successful conversational helper agent.

The main contributions are concluded as follows:
\begin{itemize}

    \item We present the Respond to Help Requests (R2H) benchmark as a test-bed for automatically evaluating the capabilities of multi-modal conversational navigation-helper agent, that helps task performers complete tasks by providing natural language responses to inquiries based on environment information. 
    \item We build two task helper agents, a novel task-oriented multi-modal helper agent, SeeRee, utilizing the Conditional Optimized Sparse (COS) attention mask and noise-free step-by-step instructions (Parse by Step) and a multi-modal LLM helper with mPLUG-Owl \cite{ye2023mplugowl}.
    \item Our experiments on the R2H benchmark and human evaluations of two helper agents over baseline also indicate that a closer linguistic resemblance to human helpers does not automatically translate into a more effective conversational helper agent.
\end{itemize}

\begin{figure*}
    \centering
        \setlength{\abovecaptionskip}{0.1cm}
    \subfloat[Benchmark for task performers.]{\includegraphics[height=6cm]{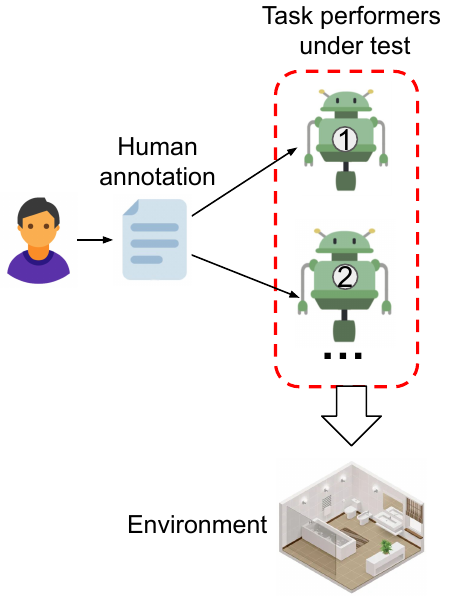}
    \label{new_f2_1}
    }
    \hspace{1cm}
    \subfloat[Benchmark for helper-performer agent pairs.]{\includegraphics[height=6cm]{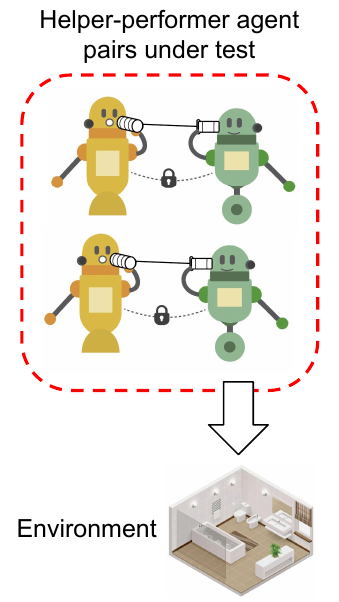}
    \label{new_f2_2}
    }
    \hspace{1cm}
    \subfloat[Benchmark for helper agents.]{\includegraphics[height=6cm]{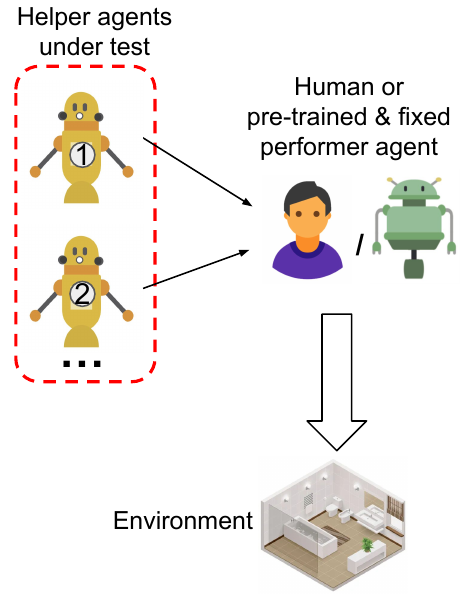}
    \label{new_f2_3}
    }
    \caption{Comparison between different dialog-based embodied benchmark types. The benchmark either (i) evaluates task performers, where performers follow instructions in human annotations, (ii) evaluates helper-performer pairs, where the helper and performer agents need to learn and be evaluated together jointly, or (iii) evaluates helper agents only (our R2H benchmark), where helpers need to provide appropriate instructions according to performer and environment information. 
    }
\end{figure*}

\section{The R2H Benchmark}

Significant differences exist between building a helper agent and developing task performer agents for dialog-based multi-modal navigation tasks. Task performers, such as ones in CVDN~\cite{thomason2020vision}, DialFRED~\cite{gao2022dialfred}, and AVDN~\cite{fan2022aerial}, are challenged as command followers, evaluated based on their performance of following dialog histories from human annotations, as shown in Figure~\ref{new_f2_1}. However, the helper agent works as a supportive role, responds to questions from the task performer to facilitate task success. Therefore, the evaluation of the helper agent’s performance could not be solely dependent on the agent itself, but on how the task performer benefited from the response. Therefore building the helper agent requires evaluations in a collaborative setting with task performers. 

Involving humans as task performers to evaluate the helper agent is ideal but expensive. Alternatively, inspired by \citet{padmakumar2022teach}, \citet{nguyen2019help} and \citet{roman2020rmm} that build helper and performer agents to collaborate as shown in Figure~\ref{new_f2_2}, we introduce the Respond to Help Requests (R2H) benchmark, involving task performer in the evaluation process as shown in Figure~\ref{new_f2_3}). In this way, the helper agent can be assessed comprehensively and realistically. R2H benchmark tests the agent's ability to respond effectively in a wide range of scenarios, including two novel tasks, the Respond to Dialog History (RDH) task and the Respond during Interaction (RdI) task, building upon three existing vision-and-dialog navigation datasets. RDH task evaluates helper agents in a situation where partial human dialog history is provided and the RdI task aims at challenging the helper agents with real collaborative scenarios.




\subsection{Respond to Dialog History Task}
task, I think that is a stand alone task.
Respond to Dialog History (RDH) Task focuses on evaluating the accuracy and completeness of the response from helper agents. The helper agent is challenged with understanding the dialog history and responds to help the task performer based on information about the task and environment in the form of image sequences. We developed environment-specific scripts to generate the image sequence which is introduced in Section \ref{adapting_dataset}. After the responses $\hat{r}_i$ generated, they will be concatenated with all available human dialog history $h_{i-1} = {\{ q_0, r_0, \dots, q_{i-1}, r_{i-1} \}}$ in the corresponding trajectory before the inquiries $q_i$ from human task performers. As a result, the generated response from the helper agent forms a new dialog history $\hat{h} = \{h_{i-1}, q_i, \hat{r}_i\}$ which becomes the input to the task performer.

\subsection{Respond during Interaction Task}

The Respond during Interaction (RdI) task challenges the ability of helper agents to cooperate consistently with the task performer. Similar to the RDH task, the RdI task involves a pre-trained task performer agent predicting navigation actions based on dialog histories. However, unlike the RDH task, no dialog history is initially provided in the RdI task, and the helper agent needs to respond to the navigation inquiries from the task performer constantly during the navigation. The task performer agent initiates inquiries $\hat{q}_i$ for help when needed and navigates based on the ongoing dialog between itself and the helper agent $\hat{h}_i = \{ \hat{q}_0, \hat{r}_0, \dots, \hat{q}_i, \hat{r}_i \}$, where $\hat{r}_i$ is the real-time responses from the helper agent to $\hat{q}_i$. The dialog $\hat{h}_i$ as a result of interaction serves as the primary source of guidance for the task performer agent, making the consistent quality of the helper agent's responses crucial for successful navigation. Additionally, since there is multi-turn helper-performer cooperation involved in the RdI task, the helper's communication efficiency can be evaluated by the number of conversation turns required for a fixed performer's outcome.

\subsection{Task Performer Agent}
Our R2H benchmark requires task performer agents to form helper-performer cooperation. In the RDH task, the task performer agent predicts navigation actions based on dialog history in specific environments. As for the RdI task, the task performer also needs to generate navigation inquiries to accomplish the navigation task better.

R2H benchmark adopts state-of-the-art open-sourced task performer agents for vision-and-language navigation datasets. The task performer agent is pre-trained on the original training set with human dialogs $h_i$ including the response from the human helper $r_i$ and predicts actions based on $\hat{h}$ for completing the task. Therefore, the progress and success made by the task performer can be seen as a test of the accuracy and completeness of the responses generated by the helper agent.




\subsection{Adapting Existing Datasets}
\label{adapting_dataset}
R2H benchmark establishes the same tasks across different datasets.
\paragraph{Datasets} R2H benchmark is built upon existing vision-and-dialog navigation datasets with dialogs between task performers and helpers.

\begin{itemize}
    \item \textbf{CVDN} \cite{thomason2020vision} is situated in the Matterport3D simulator \cite{Matterport3D} with photo-realistic scenes. The dataset records the collaboration between the human task performer and the human helper to complete navigation tasks of finding target objects in different indoor environments. 
    \item \textbf{DialFRED} \cite{gao2022dialfred} is built on Ai2-thor \cite{kolve2017ai2} simulator with synthetic views. Similar to the CVDN, the helper and task performer collaborate to navigate to targets. However, each trajectory only corresponds to one pair of inquiry and response, making it unsuitable for RdI tasks.
    \item \textbf{AVDN} \cite{fan2022aerial} is an aerial vision-and-language dataset that includes dialogs, navigation trajectories, and visual observation between the helper and task performer. The dataset is annotated upon a continuous state photo-realistic drone simulator where the goal of the task performer is to control the drone in the simulator with a top-down view to navigate to a certain destination.
\end{itemize}

\paragraph{Environment-specific Adaption} Given the variability of available information across different datasets and environments, the R2H benchmark is designed with a harmonizing approach, converting the environment-specific information into image sequences. Script-based samplers are designed for each environment to generate the image sequences by leveraging oracle information. The sampler outputs image sequences showing the task performer's views on the shortest path to the destination. Especially for the CVDN dataset, following the data collection process, a connectivity graph for viewpoints is used to generate the shortest path and therefore the image sequence length is variable but limited to views within 5 viewpoints of the current position. Examples are shown in the appendix. For the AVDN dataset, since the allocentric direction description, such as “turn south” could be involved, we keep the image sequence all oriented with the north at the top and indicate the drone direction with a red arrow. As a result, helper agents can be evaluated in different datasets with the same input format, which enhanced the benchmark's versatility to adapt further to new datasets.

\subsection{Metrics}
Since we aim to evaluate how capable the response generated by helper agents is in helping the task performer, we adopt the primary metrics for task completion from each dataset: Goal Progress (GP) in CVDN evaluates the distance of the progress made towards the destination, where it is computed as the trajectory length, deducted by the remaining  trajectory from the current location to the destination viewpoint; Success Rate (SR) in DialFRED, shows the ratio of tasks being completed successfully; Success weighted by inverse Path Length (SPL) \cite{anderson2018vision} 
as the dominant metric in AVDN measures the Success Rate weighted by the total length of the navigation trajectory.

\section{Models}
\subsection{SeeRee}
\begin{figure}[t]
    \centering
    \setlength{\abovecaptionskip}{0.1cm}
    \includegraphics[width=0.38\textwidth]{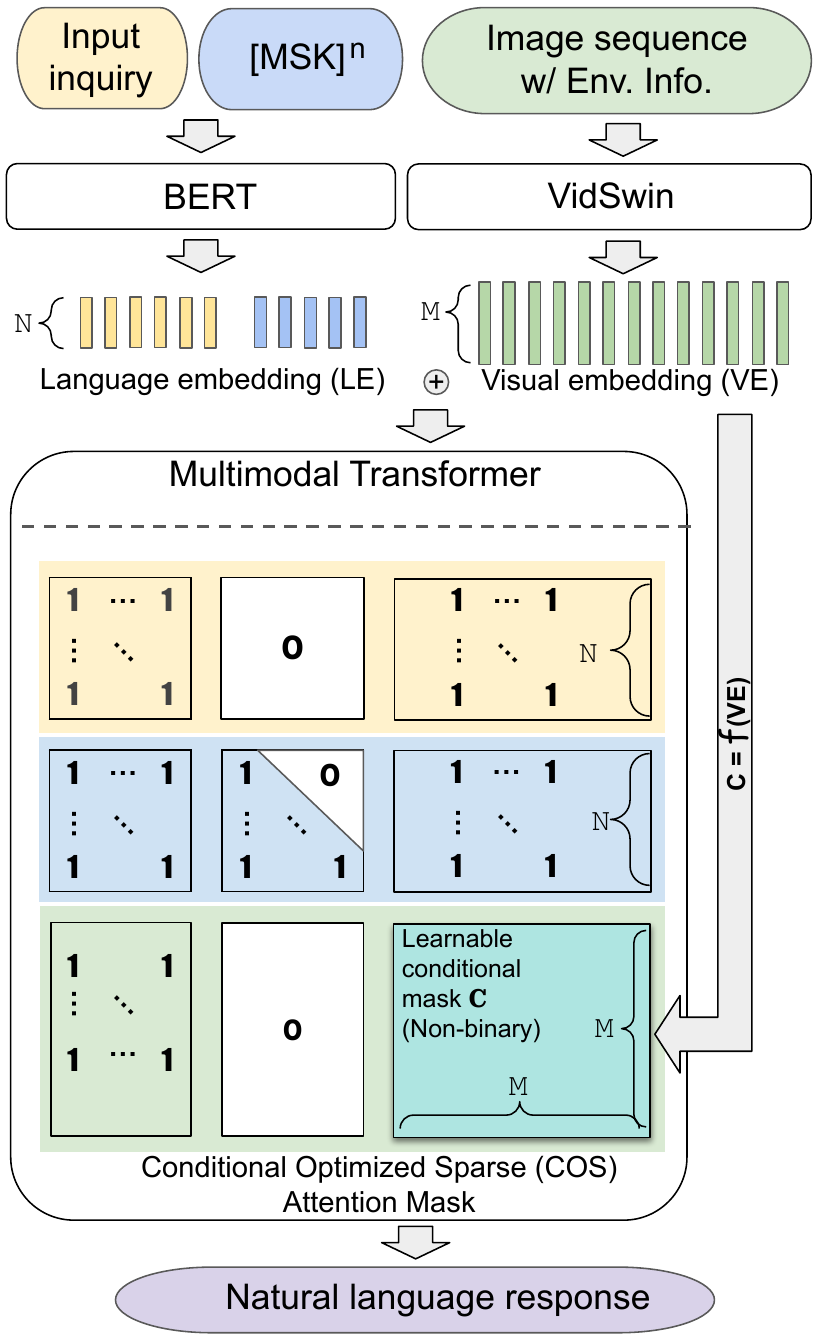}
    \caption{Overview of SeeRee. The visual and text inputs are encoded and fed into the multi-modal transformer, where Conditional Optimized Sparse (COS) attention mask is applied. The COS attention mask has fixed binary values except for the learnable conditional mask $C$ for visual embedding (VE) that is conditioned on VE itself. Yellow, blue, and green colored rows correspond to the attention masks for LE of input inquiry and generated response and VE, respectively.
    }
    \label{fig:figure2}
\end{figure}

In this section, we introduce the helper agent that can \emph{see and respond}, named SeeRee. SeeRee generates responses to task-oriented navigation inquiries from the task performer. 
As is illustrated in Figure~\ref{fig:figure2}, our helper agent SeeRee generates natural language responses to the task performer's inquiries based on the task and environment information that is not aware by the task performer. The image sequences are padded to a fixed length \cite{lin2022swinbert}, encoded by Video Swin Transformer \cite{liu2022video} and then concatenated with BERT text embeddings \cite{devlin2018bert}. Finally, the embeddings are fed into a multi-modal transformer which generates the natural language response in an auto-regressive way. SeeRee is trained end-to-end with Mask Language Modeling \cite{devlin2018bert} and please refer to the appendix for the training detail.

\subsubsection{Multi-modal Transformer}
Following prior multi-modal language generation studies \cite{hu2020vivo, lin2022swinbert}, our multi-modal transformer takes as input the embedding containing both text and image information and generates natural language responses in a unidirectional sequence-to-sequence generation process. We treat the input inquiries as prompts for generating the response, and
a special token [CLS] is added to the end of the inquiry. 

At inference time, the text is generated in an auto-regressive manner, where we insert multiple [MSK] tokens after the [CLS] token and predict tokens to replace [MSK]tokens one by one unidirectionally until the prediction is [EOS] or all the [MSK] tokens are predicted. 
\subsubsection{Conditional Optimized Sparse (COS) Attention Mask}
\label{section:3.cos mas}
One challenge in generating responses for dialog-based embodied  tasks is effectively modeling the long input image sequence, reducing the redundancy in the repetitive images but keeping the critical details. To this end, we introduce a Conditional Optimized Sparse (COS) attention mask for the multi-modal transformer, as shown in Figure~\ref{fig:figure2}. The mask can be divided into three row sections, corresponding to in what range the embeddings for input inquiry, the response generated, input image sequence can attend to, respectively. The first row section shows that the language embedding (LE) for input inquiry can attend to itself and the input visual embedding (VE). The second row section means the LE for the generated response can attend to the VE and all the previous LE, allowing unidirectional text generation \cite{devlin2018bert}. The third row section indicates that the VE can attend to partial itself and the LE of input inquiry. Especially, instead of being fulling binary and pre-defined, a learnable conditional mask $C$ that is non-binary and sparse is adopted in the third row section of the COS attention mask, controlling the self-attention of the VE. $C$ is conditioned on VE, and the mapping is modeled by:
\begin{equation}
    C = \sigma ( f(\text{VE})),
\end{equation}

where  $\sigma(x) = \frac{1}{1+e^{-x}}$ and $f$ is a multi-layer perceptron. In this way, our COS attention mask uses a conditional optimizing strategy that optimizes the attention mask based on the image sequence. 
As a result, COS attention mask enables better encoding and understanding of long visual input and improves the response generation result.


\subsubsection{Response Preprocessing with Parse by Step}
\label{section:3.pbs}
Training a helper agent to imitate the responses in the human dialog directly may not be optimal as they may be unorganized and include utterances irrelevant to the task. 
Structured step-by-step instructions are easier for the helper agent to learn.
Therefore, inspired by the idea of in-context learning \cite{brown2020language,kojima2022large}, we propose a Parse by Step method that prompts GPT-3 \cite{brown2020language} to preprocess the ground-truth responses in the training data. We detail the designed prompts and examples in the appendix.
After Prase by Step, the preprocessed training data is parsed in a step-by-step manner with a streamlined language pattern.
As a result, the learning objectives of SeeRee is the preprocessed human response $Y = P(R)$, where $P$ is the Parse by Step method, and $R$ is the original human response in the dialog of the training set.

\subsection{Multi-modal Large Language Model}
Besides our SeeRee model, we introduce another navigation-helper agent constructed from a multi-modal large language model (LLM). Specifically, we employ mPLUG-Owl \cite{ye2023mplugowl}, a State-Of-The-Art multi-modal LLM that is able to take as input an image sequence and text for language generation. mPLUG-Owl is originally trained with a large amount of uni-modal and multi-modal data in two stage paradigm including instruction tuning. To leverage the extensive knowledge that mPLUG-Owl accumulated through training and avoid the potential issue associated with limited task-specific training data, we adopt mPLUG-Owl in a zero-shot manner. The build of a helper agent based on mPLUG-Owl serves as a valuable comparison for SeeRee and sheds light on the potential of leveraging LLMs in building navigation-helper agents.

\section{Experiments}




\begin{table}[t]
\setlength{\abovecaptionskip}{0.1cm}
    \centering

\begin{subtable}[t]{0.48\textwidth}
\resizebox{\columnwidth}{!}{
\setlength\tabcolsep{0pt}
\begin{tabular}{ll|cc|cc|cc} 
\toprule
& & \multicolumn{2}{c|}{\textbf{CVDN}} & \multicolumn{2}{c|}{\textbf{DialFRED }} &\multicolumn{2}{c}{\textbf{AVDN}} \\
\cmidrule(lr){3-4} \cmidrule(lr){5-6} \cmidrule(lr){7-8} 

&\multicolumn{1}{c|}{\textbf{Helper}}
&  \multicolumn{1}{l}{$\begin{array}{c}
     \textbf{Seen}  \\
     \textbf{GP}\uparrow
\end{array}$} 
&  \multicolumn{1}{l|}{$\begin{array}{c}
     \textbf{Unseen}  \\
     \textbf{GP}\uparrow
\end{array}$} 
&  \multicolumn{1}{l}{$\begin{array}{c}
     \textbf{Seen}  \\
     \textbf{SR}\uparrow
\end{array}$} 
&  \multicolumn{1}{l|}{$\begin{array}{c}
     \textbf{Unseen}  \\
     \textbf{SR}\uparrow
\end{array}$} 
&  \multicolumn{1}{l}{$\begin{array}{c}
     \textbf{Seen}  \\
     \textbf{SPL}\uparrow
\end{array}$} 
&  \multicolumn{1}{l}{$\begin{array}{c}
     \textbf{Unseen}  \\
     \textbf{SPL}\uparrow
\end{array}$} 
\\
\midrule
\multicolumn{2}{l|}{$\begin{array}{l}
    \text{Human Annotator}\\
\end{array}$} & 6.9 & 5.1 & 49.1 & 33.4 & 14.7 & 16.5  \\
\midrule

 \multicolumn{2}{l|}{$\begin{array}{c}
    \text{RMM }\mathcal{G}
\end{array}$} & 4.7 & 2.8 & 46.5 & 32.1 & 2.4 & 3.6\\

 \multicolumn{2}{l|}{$\begin{array}{c}
    \text{Multimodal LLM}
\end{array}$} & 5.3 & 3.6 & 47.0 & \textbf{33.8} & 0.7 & 1.5 \\

 \multicolumn{2}{l|}{$\begin{array}{l}
    \text{SeeRee}\\
\end{array}$}  & \textbf{6.5} & \textbf{4.9} & \textbf{49.1} & 33.1 & \textbf{4.6} & \textbf{4.4} \\

\bottomrule
\end{tabular}
}
\end{subtable}

\caption{Results of Respond to Dialog History (RDH) task on CVDN, DialFRED, and AVDN dataset \cite{thomason2020vision,gao2022dialfred,fan2022aerial}. We replace the original response in validation sets with responses from different helper agents. With a fixed task performer agent, a better performance of the performer agent represents a more effective response from the helper agent.}
\label{tab:results_RDH}
\end{table}

\label{R2HexperimentSec}

\subsection{Setup}
We initialize the encoders and multi-modal transformers in SeeRee with weights from SwinBert \cite{lin2022swinbert} to benefit from its image sequence understanding ability and then finetune SeeRee on the training set of each dataset separately. 
We adopt the RMM Guide model ($\mathcal{G}$) \cite{roman2020rmm} as the baseline model in our experiments. Please refer to the appendix for more implementation details. 

\subsection{RDH Task}

\paragraph{Task Performer Agents}
For RDH task on CVDN dataset \cite{thomason2020vision}, we use HAMT\footnote{\footnotesize{\url{https://github.com/cshizhe/VLN-HAMT}.}} \cite{chen2021hamt}, pre-trained on the original Navigation from Dialog History (NDH) task in CVDN as the task performer agent. 
For AVDN \cite{fan2022aerial} and DialFRED \cite{gao2022dialfred} datasets, we leverage the task performer in the original work, i.e., HAA-transformer and DialFRED model \footnote{Due to unavailable model weights, we reproduced the model with comparable results as in the original paper.}. We further detail the task performers in the appendix. 

\paragraph{Result}
As indicated in Table~\ref{tab:results_RDH}, the task performer agent attains the best overall performance across all three datasets when guided by responses generated by the SeeRee helper agent. On both CVDN and DiaFRED datasets, SeeRee demonstrates performance levels that are strikingly similar to those of human helpers.
This is likely attributable to the fact that the environments used in the CVDN are visually akin to the data employed during the pre-training phase of SeeRee's encoders. Such similarity facilitates efficient fine-tuning and leads to heightened performance. In addition, for the DialFRED dataset, due to the predominance of template-based human utterances, the challenge in response generation is simplified. 
Meanwhile, the AVDN dataset is more challenging due to the drone simulator's adoption of a continuous control space, and complex visual information leads to difficulty in reaching the precise destination. Still, SeeRee outperforms the baseline by a large margin. We present results from more LLM baselines and case studies of the RDH task in Appendix~\ref{add_LLM_Baseline} and \ref{case}. 


\subsection{RdI Task}

\paragraph{Task Performer Agents}
We utilize the task performer agent from the pioneering work RMM \cite{roman2020rmm} deployed on CVDN dataset \cite{thomason2020vision}, which combines two components: the LSTM-based RMM Navigator model ($\mathcal{N}$), responsible for predicting navigation actions, and RMM Questioner model ($\mathcal{Q}$), generating inquires to the helper agent after every four navigation actions. Despite its comparatively simple structure, the RMM task performer stands out as the sole task performer developed that possesses both language and navigation capabilities: generating navigation actions and crafting natural language navigation inquiries. Future works for building such language- and navigation-capable task performers in AVDN and DialFRED environments are needed.
Furthermore, given the flexibility of our system, should a superior model be developed, the task performer could be effortlessly swapped.

\begin{figure}[t]
    \centering
        \setlength{\abovecaptionskip}{0.1cm}
    
    \includegraphics[width=0.48\textwidth]{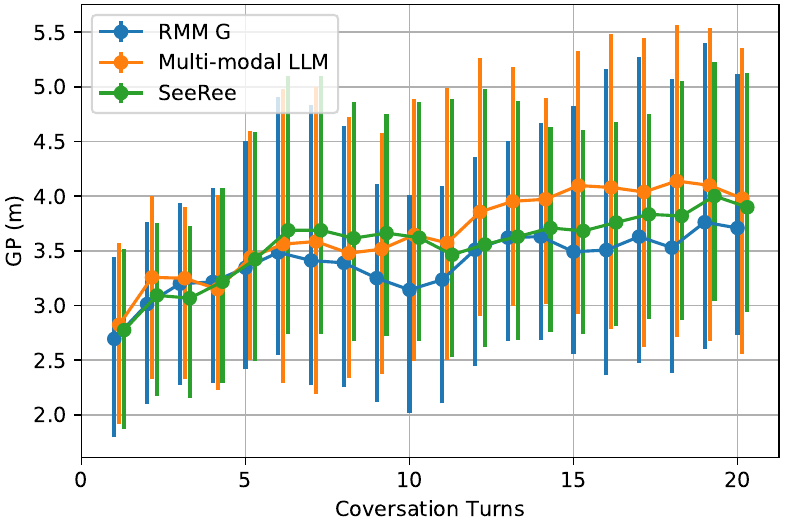}
    \caption{Results of Respond during Interaction (RdI) task on CVDN dataset \cite{thomason2020vision}. The mean Goal Progress (GP) of the same task performer collaborating with different helper agents is plotted with respect to the number of conversation turns happened. Error bars show the relative range of the GP. Multi-modal LLM enables a better but noisier performance of the task performer than SeeRee. \label{RdI_cvdn}}
\end{figure}

\paragraph{Result}
The RdI task is conducted with a maximum 20 turns of conversation between the task performer and helper on the unseen validation set of the CVDN dataset. As depicted in Figure~\ref{RdI_cvdn}, we plot the mean GP of the task performer agent concerning the dialog turns happened during the navigation. As the dialog turn between helper and performer increases chronologically, more information is provided to the task performer. The navigation stops once the task performer reaches the maximum steps, 20, or a stop action is generated where the goal progress remains constant thereafter. Based on the result, the multi-modal LLM as the helper agent facilitates the most effective performance from the task performer at the end of the maximum conversation turns. However, it's worth noting that the response from SeeRee is shown to be less noisy on its effect and therefore is potentially more reliable over extended dialogue turns whereas multi-modal LLM tends to generate responses with a more varied effectiveness. Moreover, with less than half the maximum dialogue turns, SeeRee's responses yield a superior result that is 90\% close to the navigation result at the end of the maximum turn. This underscores SeeRee's greater communication efficiency.

\subsection{Ablation Study}

\begin{table}[t]
 \centering
 \setlength{\abovecaptionskip}{0.1cm}
  \setlength\tabcolsep{2pt}
\resizebox{\linewidth}{!}{
\begin{tabular}{l|cc|ccc|ccc} 
\toprule
\multicolumn{1}{c|}{\multirow{2}{*}{\textbf{Helper}}}  & \multirow{2}{*}{$\begin{array}{c} \textbf{COS}\\ \textbf{attention mask}\end{array}$} & \multirow{2}{*}{$\begin{array}{c} \textbf{Parse by Step}\end{array}$} & \multicolumn{3}{c|}{$\begin{array}{c} \textbf{Seen}\\ \textbf{Validation}\end{array}$} & \multicolumn{3}{c}{$\begin{array}{c} \textbf{Unseen}\\ \textbf{Validation}\end{array}$}  \\ 
\cmidrule{4-9}
\multicolumn{1}{c|}{} &&  & GP& B2 & R  & GP& B2 & R \\ 
\cmidrule{1-3}\cmidrule{4-9}
\multirow{2}{*}{$\begin{array}{l} \text{Human}\\ \text{Annotator} \\\end{array}$}  & - & \xmark & 6.9 & - & - & 5.1 & - & -\\
$\begin{array}{l} \text{}\\\end{array}$ & - & \cmark & 7.3 & -  & -  & 5.1 & -  & - \\ 
\cmidrule{1-4}\cmidrule{5-6}\cmidrule{7-7}\cmidrule{8-9}
\multirow{3}{*}{$\begin{array}{l} \text{SeeRee}\\\end{array}$} & \xmark & \xmark & 5.4 & 9.9 & 21.8 & 4.7 & 10.0 & 22.1\\
 & \cmark & \xmark & 5.9 & \textbf{14.4} & \textbf{25.5} & 4.7 & \textbf{13.9} & \textbf{25.5}\\
 & \cmark & \cmark & \textbf{6.5} & 13.8 & 24.2 & \textbf{4.9} & 13.2 & 24.1\\
\bottomrule
\end{tabular}
}
\caption{Ablation study of the SeeRee agent on CVDN dataset \cite{thomason2020vision} based on RDH task GP and language generation metrics, BLUE2 (B2) and ROUGE-L (R). We directly apply our Parse by Step method to the original validation data created by humans, showing that Parse by Step maintains essential task-related information. The results show the effectiveness of COS attention mask and Parse by Step in task-oriented response generation.  \label{tab:ablation}
}
\end{table}

\begin{table}[t]
    \centering
\setlength{\abovecaptionskip}{0.1cm}
\setlength\tabcolsep{0pt}
\resizebox{\columnwidth}{!}{
\begin{tabular}{llc|c|cc} 
\toprule
\multicolumn{3}{c|}{}& \multicolumn{1}{c|}{$\begin{array}{c}
     \textbf{Task}  \\
     \textbf{Completion}
\end{array}$} & \multicolumn{2}{c}{$\begin{array}{c}
     \textbf{Subjective}  \\
     \textbf{Response Evaluation}
\end{array}$}\\
 \cmidrule(lr){4-4} 
 \cmidrule(lr){5-6} 

\multicolumn{2}{c|}{\textbf{Helper}} &
\multicolumn{1}{c|}{$\begin{array}{c}
     \textbf{Task}\\
     \textbf{Performer}
\end{array}$}
& \multicolumn{1}{c|}{$\begin{array}{c}
     \text{GP}\uparrow
\end{array}$}
& \multicolumn{1}{c}{$\begin{array}{c}
     \text{Naturalness}\uparrow
\end{array}$} & \multicolumn{1}{c}{$\begin{array}{c}
     \text{Faithfulness}\uparrow
\end{array}$}\\
\midrule
\multicolumn{2}{l|}{$\begin{array}{l}
    \text{No helper}\\
\end{array}$} & \multirow{4}{*}{$\begin{array}{l}
    \text{Human}
\end{array}$}  & 3.54 & - & -  \\

 \multicolumn{2}{l|}{$\begin{array}{c}
    \text{RMM }\mathcal{G}
\end{array}$} &  & 6.32 & 73/100 & 59/100  \\
\multicolumn{2}{l|}{$\begin{array}{l}
    \text{Multi-modal LLM}\\
\end{array}$} &  &  8.44 &  \textbf{75/100} & 60/100 \\

 \multicolumn{2}{l|}{$\begin{array}{l}
    \text{SeeRee}\\
\end{array}$} &  &  \textbf{9.89} &  72/100 & \textbf{75/100}  \\

\bottomrule
\end{tabular}
}
\caption{Results of human evaluation on RdI task on the CVDN \cite{thomason2020vision}. The human tester plays the task performer role via an interface interacting with the helper agents. Task completion and subjective response evaluation are collected. The response from SeeRee is most effective despite being less natural. 
 \label{tab:table_2} }
\end{table}

Based on CVDN dataset \cite{thomason2020vision}, we conduct two ablation studies to explore the Parse by Step method and COS attention mask for SeeRee.
We first let the task performer infers on the human dialog from the original dataset, but the response within is processed by our Parse by Step method. Then we create ablated SeeRee model by repeating the same training on SeeRee model twice, but incrementally replacing a naive fixed attention mask with the COS attention mask and applying our Parse by Step method.

\paragraph{Response Effectiveness Result}
The first ablation study that evaluate of Parse by Step method on the original validation set serves as a sanity check. As the result in Table~\ref{tab:ablation}, it shows that human responses processed with Pase by Step keep the information contained in the original response and they are overall equally capable as the original response. Additionally in the second ablation study, the GP of the same task performer cooperated with different ablated SeeRee shows that both COS attention mask and Parse by Step leads to major improvements to the effectiveness of the response in facilitating task success. 




\paragraph{Language Generation Similarity Result} As shown in Table~\ref{tab:ablation}, we also evaluate the ablated models with language generation metrics, BLUE2 \cite{papineni2002bleu} and ROUGE-L \cite{lin2004automatic}. BLUE2 score and ROUGH-L score drops when Parse by Stop method is applied, but GP in the RDH task receives a major increase. This indicates that a high language similarity to human responses does not necessarily equate to a more effective conversational helper agent.

\subsection{Human Evaluation}
To further evaluate the performance of helper agents, we conduct human evaluations based on the RdI task. Human participants act as task performers navigating 60 randomly selected trajectories in validation sets of CVDN \cite{thomason2020vision}. During the simulation, participants can control their movement in the environment using their keyboards and ask questions to the helper agent whenever needed. We evaluate both settings where the helper agent exists or not. Human participants are also asked to rate the naturalness and faithfulness of the response from the helper. The average GP and subjective rating of the responses are shown in Table \ref{tab:table_2}. The result shows that SeeRee provides the best results in terms of task completion. Through subjective evaluation, we find that SeeRee achieves significantly higher scores in terms of response faithfulness. Despite being trained with data preprocessed by Parse by Step, where the training supervision is no longer the original human utterances, it still achieves a close score of naturalness compared to the baseline model trained on original human responses. Through these evaluations, we show the ability of SeeRee to help human users complete embodied tasks.

\section{Related Work}

\paragraph{Dialog-based Multi-modal Embodied Benchmarks}

Previous dialog-based multi-modal embodied benchmarks usually focus on evaluating either task performers~\cite{thomason2020vision, gao2022dialfred, shi2022learning, gu2022vision} or the corresponding pairs of task performer and helper~\cite{roman2020rmm, hahn2020you, padmakumar2022teach}. For instance, the CVDN \cite{thomason2020vision} evaluates a task performer to navigate to a desired room by dialog histories. \citet{gao2022dialfred} developed a dialogue-enabled embodied instruction following benchmark, DialFRED, based on the ALFRED benchmark~\cite{ALFRED20} and presented a task performer framework. Further, there is a wide range of activities studied in these tasks, such as navigating to a specific location~\cite{roman2020rmm}, locating positions~\cite{hahn2020you}, and interacting objects~\cite{padmakumar2022teach}. Compared to these benchmarks, our R2H benchmark aims for better helper agents and is the only benchmark for sole helper evaluation.

\paragraph{Multimodal-based Language Generation}
Building helper agents is in line with a growing collection of methods applied to the visual question-answering task. 
A unified vision-and-language framework can be trained to handle a variety of tasks, including the question-answering problem, using a single objective~\cite{cho2021unifying, wang2022git}.
\citet{fu2021violet} tackled the video question-answering problem by adopting a video transformer to model the temporal dynamics of video inputs explicitly. One problem shared by these works is the input frames to the model are limited in quantity, whereas the helper agent has to take a long image sequence as input to include adequate information. \citet{lin2022swinbert} developed a learnable sparse attention mask for video caption that enables long frame sequence as input. However, the learned sparse attention mask is fixed after training, lacking generalization ability compared to the COS attention mask of SeeRee.

\section{Conclusion}

In this paper, we introduce the Respond to Help Requests (R2H) benchmark with two tasks, Respond to Dialog History (RDH) and Respond during Interaction (RdI), assessing a helper agent's guidance capabilities. With the R2H benchmark, we build and evaluate two navigation-helper agents, SeeRee and Multi-modal LLM model. The results show that they both outperformed the baseline model. Through further ablation study and human evaluation, we prove the effectiveness of SeeRee in assisting humans with tasks and argue that the effectiveness of responses cannot be determined by their linguistic resemblance to human dialogue alone.

\section*{Limitations}

R2H benchmark presents a platform for helper agents that create natural language to address queries from task performers. Yet, the assessment of such an helper agent mandates a capable task performer agent that can not only navigate but also communicate, given that the efficacy of the helper can be gauged through final task fulfillment, and the task performer shouldn't act as a constraint. Also, the complexity of the real world surpasses that of a simulated environment, thereby imposing additional prerequisites for the task performer.
Furthermore, the lack of abundant dialog-based embodied datasets also restricts the progression of both performer and helper agents.

\section*{Ethics Statement}
The human evaluation part of this project is classified as exempt by Human Subject Committee vis IRB protocols. Additionally, we recognize the potential ethical issues related to training language generation models using human dialog data, such as the possibility of the model learning harmful language without understanding its context. To mitigate this concern, our proposed data preprocessing method, Parse by Step, converts the responses in the training data into more structured and task-specific instructions, effectively reducing the presence of noise and profanity in the training data. As a result, the likelihood of our model generating inappropriate responses is greatly minimized.

\bibliography{custom}

\begin{thebibliography}{29}
\expandafter\ifx\csname natexlab\endcsname\relax\def\natexlab#1{#1}\fi

\bibitem[{Anderson et~al.(2018)Anderson, Wu, Teney, Bruce, Johnson,
  S{\"u}nderhauf, Reid, Gould, and Van Den~Hengel}]{anderson2018vision}
Peter Anderson, Qi~Wu, Damien Teney, Jake Bruce, Mark Johnson, Niko
  S{\"u}nderhauf, Ian Reid, Stephen Gould, and Anton Van Den~Hengel. 2018.
\newblock Vision-and-language navigation: Interpreting visually-grounded
  navigation instructions in real environments.
\newblock In \emph{Proceedings of the IEEE conference on computer vision and
  pattern recognition}, pages 3674--3683.

\bibitem[{Brown et~al.(2020)Brown, Mann, Ryder, Subbiah, Kaplan, Dhariwal,
  Neelakantan, Shyam, Sastry, Askell et~al.}]{brown2020language}
Tom Brown, Benjamin Mann, Nick Ryder, Melanie Subbiah, Jared~D Kaplan, Prafulla
  Dhariwal, Arvind Neelakantan, Pranav Shyam, Girish Sastry, Amanda Askell,
  et~al. 2020.
\newblock Language models are few-shot learners.
\newblock \emph{Advances in neural information processing systems},
  33:1877--1901.

\bibitem[{Chang et~al.(2017)Chang, Dai, Funkhouser, Halber, Niessner, Savva,
  Song, Zeng, and Zhang}]{Matterport3D}
Angel Chang, Angela Dai, Thomas Funkhouser, Maciej Halber, Matthias Niessner,
  Manolis Savva, Shuran Song, Andy Zeng, and Yinda Zhang. 2017.
\newblock Matterport3d: Learning from {RGB-D} data in indoor environments.
\newblock \emph{International Conference on 3D Vision (3DV)}.

\bibitem[{Chen et~al.(2021)Chen, Guhur, Schmid, and Laptev}]{chen2021hamt}
Shizhe Chen, Pierre-Louis Guhur, Cordelia Schmid, and Ivan Laptev. 2021.
\newblock History aware multimodal transformer for vision-and-language
  navigation.
\newblock In \emph{NeurIPS}.

\bibitem[{Cho et~al.(2021)Cho, Lei, Tan, and Bansal}]{cho2021unifying}
Jaemin Cho, Jie Lei, Hao Tan, and Mohit Bansal. 2021.
\newblock Unifying vision-and-language tasks via text generation.
\newblock In \emph{International Conference on Machine Learning}, pages
  1931--1942. PMLR.

\bibitem[{Devlin et~al.(2019)Devlin, Chang, Lee, and
  Toutanova}]{devlin2018bert}
Jacob Devlin, Ming-Wei Chang, Kenton Lee, and Kristina Toutanova. 2019.
\newblock Bert: Pre-training of deep bidirectional transformers for language
  understanding.

\bibitem[{Fan et~al.(2022)Fan, Chen, Jiang, Zhou, Zhang, and
  Wang}]{fan2022aerial}
Yue Fan, Winson Chen, Tongzhou Jiang, Chun Zhou, Yi~Zhang, and Xin~Eric Wang.
  2022.
\newblock Aerial vision-and-dialog navigation.
\newblock \emph{arXiv preprint arXiv:2205.12219}.

\bibitem[{Fu et~al.(2021)Fu, Li, Gan, Lin, Wang, Wang, and Liu}]{fu2021violet}
Tsu-Jui Fu, Linjie Li, Zhe Gan, Kevin Lin, William~Yang Wang, Lijuan Wang, and
  Zicheng Liu. 2021.
\newblock Violet: End-to-end video-language transformers with masked
  visual-token modeling.
\newblock \emph{arXiv preprint arXiv:2111.12681}.

\bibitem[{Gao et~al.(2022)Gao, Gao, Gong, Lin, Thattai, and
  Sukhatme}]{gao2022dialfred}
Xiaofeng Gao, Qiaozi Gao, Ran Gong, Kaixiang Lin, Govind Thattai, and Gaurav~S
  Sukhatme. 2022.
\newblock Dialfred: Dialogue-enabled agents for embodied instruction following.
\newblock \emph{arXiv preprint arXiv:2202.13330}.

\bibitem[{Gu et~al.(2022)Gu, Stefani, Wu, Thomason, and Wang}]{gu2022vision}
Jing Gu, Eliana Stefani, Qi~Wu, Jesse Thomason, and Xin Wang. 2022.
\newblock Vision-and-language navigation: A survey of tasks, methods, and
  future directions.
\newblock In \emph{Proceedings of the 60th Annual Meeting of the Association
  for Computational Linguistics (Volume 1: Long Papers)}, pages 7606--7623.

\bibitem[{Hahn et~al.(2020)Hahn, Krantz, Batra, Parikh, Rehg, Lee, and
  Anderson}]{hahn2020you}
Meera Hahn, Jacob Krantz, Dhruv Batra, Devi Parikh, James Rehg, Stefan Lee, and
  Peter Anderson. 2020.
\newblock Where are you? localization from embodied dialog.
\newblock In \emph{Proceedings of the 2020 Conference on Empirical Methods in
  Natural Language Processing (EMNLP)}, pages 806--822.

\bibitem[{Hu et~al.(2020)Hu, Yin, Lin, Wang, Zhang, Gao, and Liu}]{hu2020vivo}
Xiaowei Hu, Xi~Yin, Kevin Lin, Lijuan Wang, Lei Zhang, Jianfeng Gao, and
  Zicheng Liu. 2020.
\newblock Vivo: Surpassing human performance in novel object captioning with
  visual vocabulary pre-training.

\bibitem[{Kojima et~al.()Kojima, Gu, Reid, Matsuo, and
  Iwasawa}]{kojima2022large}
Takeshi Kojima, Shixiang~Shane Gu, Machel Reid, Yutaka Matsuo, and Yusuke
  Iwasawa.
\newblock Large language models are zero-shot reasoners.
\newblock In \emph{Advances in Neural Information Processing Systems}.

\bibitem[{Kolve et~al.(2017)Kolve, Mottaghi, Han, VanderBilt, Weihs, Herrasti,
  Gordon, Zhu, Gupta, and Farhadi}]{kolve2017ai2}
Eric Kolve, Roozbeh Mottaghi, Winson Han, Eli VanderBilt, Luca Weihs, Alvaro
  Herrasti, Daniel Gordon, Yuke Zhu, Abhinav Gupta, and Ali Farhadi. 2017.
\newblock Ai2-thor: An interactive 3d environment for visual ai.
\newblock \emph{arXiv preprint arXiv:1712.05474}.

\bibitem[{Li et~al.(2023)Li, Li, Savarese, and Hoi}]{li2023blip}
Junnan Li, Dongxu Li, Silvio Savarese, and Steven Hoi. 2023.
\newblock Blip-2: Bootstrapping language-image pre-training with frozen image
  encoders and large language models.
\newblock \emph{arXiv preprint arXiv:2301.12597}.

\bibitem[{Lin and Och(2004)}]{lin2004automatic}
Chin-Yew Lin and Franz~Josef Och. 2004.
\newblock Automatic evaluation of machine translation quality using longest
  common subsequence and skip-bigram statistics.
\newblock In \emph{Proceedings of the 42nd Annual Meeting of the Association
  for Computational Linguistics (ACL-04)}, pages 605--612.

\bibitem[{Lin et~al.(2022)Lin, Li, Lin, Ahmed, Gan, Liu, Lu, and
  Wang}]{lin2022swinbert}
Kevin Lin, Linjie Li, Chung-Ching Lin, Faisal Ahmed, Zhe Gan, Zicheng Liu,
  Yumao Lu, and Lijuan Wang. 2022.
\newblock Swinbert: End-to-end transformers with sparse attention for video
  captioning.
\newblock In \emph{Proceedings of the IEEE/CVF Conference on Computer Vision
  and Pattern Recognition}, pages 17949--17958.

\bibitem[{Liu et~al.(2022)Liu, Ning, Cao, Wei, Zhang, Lin, and
  Hu}]{liu2022video}
Ze~Liu, Jia Ning, Yue Cao, Yixuan Wei, Zheng Zhang, Stephen Lin, and Han Hu.
  2022.
\newblock Video swin transformer.
\newblock In \emph{Proceedings of the IEEE/CVF Conference on Computer Vision
  and Pattern Recognition}, pages 3202--3211.

\bibitem[{Loshchilov and Hutter(2018)}]{loshchilov2018decoupled}
Ilya Loshchilov and Frank Hutter. 2018.
\newblock Decoupled weight decay regularization.
\newblock In \emph{International Conference on Learning Representations}.

\bibitem[{Nguyen and Daum{\'e}~III(2019)}]{nguyen2019help}
Khanh Nguyen and Hal Daum{\'e}~III. 2019.
\newblock Help, anna! visual navigation with natural multimodal assistance via
  retrospective curiosity-encouraging imitation learning.
\newblock \emph{arXiv preprint arXiv:1909.01871}.

\bibitem[{Padmakumar et~al.(2022)Padmakumar, Thomason, Shrivastava, Lange,
  Narayan-Chen, Gella, Piramuthu, Tur, and Hakkani-Tur}]{padmakumar2022teach}
Aishwarya Padmakumar, Jesse Thomason, Ayush Shrivastava, Patrick Lange, Anjali
  Narayan-Chen, Spandana Gella, Robinson Piramuthu, Gokhan Tur, and Dilek
  Hakkani-Tur. 2022.
\newblock Teach: Task-driven embodied agents that chat.
\newblock In \emph{Proceedings of the AAAI Conference on Artificial
  Intelligence}, volume~36, pages 2017--2025.

\bibitem[{Papineni et~al.(2002)Papineni, Roukos, Ward, and
  Zhu}]{papineni2002bleu}
Kishore Papineni, Salim Roukos, Todd Ward, and Wei-Jing Zhu. 2002.
\newblock Bleu: a method for automatic evaluation of machine translation.
\newblock In \emph{Proceedings of the 40th annual meeting of the Association
  for Computational Linguistics}, pages 311--318.

\bibitem[{Pashevich et~al.(2021)Pashevich, Schmid, and
  Sun}]{pashevich2021episodic}
Alexander Pashevich, Cordelia Schmid, and Chen Sun. 2021.
\newblock Episodic transformer for vision-and-language navigation.
\newblock In \emph{Proceedings of the IEEE/CVF International Conference on
  Computer Vision}, pages 15942--15952.

\bibitem[{Roman et~al.(2020)Roman, Bisk, Thomason, Celikyilmaz, and
  Gao}]{roman2020rmm}
Homero~Roman Roman, Yonatan Bisk, Jesse Thomason, Asli Celikyilmaz, and
  Jianfeng Gao. 2020.
\newblock Rmm: A recursive mental model for dialogue navigation.
\newblock In \emph{Findings of the Association for Computational Linguistics:
  EMNLP 2020}, pages 1732--1745.

\bibitem[{Shi et~al.(2022)Shi, Feng, and Lipani}]{shi2022learning}
Zhengxiang Shi, Yue Feng, and Aldo Lipani. 2022.
\newblock Learning to execute actions or ask clarification questions.
\newblock In \emph{Findings of the Association for Computational Linguistics:
  NAACL 2022}, pages 2060--2070.

\bibitem[{Shridhar et~al.(2020)Shridhar, Thomason, Gordon, Bisk, Han, Mottaghi,
  Zettlemoyer, and Fox}]{ALFRED20}
Mohit Shridhar, Jesse Thomason, Daniel Gordon, Yonatan Bisk, Winson Han,
  Roozbeh Mottaghi, Luke Zettlemoyer, and Dieter Fox. 2020.
\newblock Alfred: A benchmark for interpreting grounded instructions for
  everyday tasks.
\newblock In \emph{Proceedings of the IEEE/CVF conference on computer vision
  and pattern recognition}, pages 10740--10749.

\bibitem[{Thomason et~al.(2020)Thomason, Murray, Cakmak, and
  Zettlemoyer}]{thomason2020vision}
Jesse Thomason, Michael Murray, Maya Cakmak, and Luke Zettlemoyer. 2020.
\newblock Vision-and-dialog navigation.
\newblock In \emph{Conference on Robot Learning}, pages 394--406. PMLR.

\bibitem[{Wang et~al.(2022)Wang, Yang, Hu, Li, Lin, Gan, Liu, Liu, and
  Wang}]{wang2022git}
Jianfeng Wang, Zhengyuan Yang, Xiaowei Hu, Linjie Li, Kevin Lin, Zhe Gan,
  Zicheng Liu, Ce~Liu, and Lijuan Wang. 2022.
\newblock Git: A generative image-to-text transformer for vision and language.
\newblock \emph{arXiv preprint arXiv:2205.14100}.

\bibitem[{Ye et~al.(2023)Ye, Xu, Xu, Ye, Yan, Zhou, Wang, Hu, Shi, Shi
  et~al.}]{ye2023mplugowl}
Qinghao Ye, Haiyang Xu, Guohai Xu, Jiabo Ye, Ming Yan, Yiyang Zhou, Junyang
  Wang, Anwen Hu, Pengcheng Shi, Yaya Shi, et~al. 2023.
\newblock mplug-owl: Modularization empowers large language models with
  multimodality.
\newblock \emph{arXiv preprint arXiv:2304.14178}.

\end{thebibliography}
\bibliographystyle{acl_natbib}

\appendix
\newpage

\begin{table*}[t]
    \centering
    \setlength{\abovecaptionskip}{0.1cm}
    \resizebox{\textwidth}{!}{
    \begin{tabular}{llll}
    \toprule
        \multicolumn{4}{m{24cm}}{
          $\begin{array}{rl}
              \textbf{Prompt for CVDN:} & \text{According to the given images, which way to go? Please only tell me the moving direction. Do not introduce the room details.}  \\
              & \text{{\$QUESTION} }
          \end{array}$
       }\\
           \toprule
        \multicolumn{4}{m{24cm}}{
          $\begin{array}{rl}
              \textbf{Prompt for DialFRED:} & \text{The images first show a sequence of first person view of a robot. Based on the image sequence, please answer the question  }\\
              & \text{about {\$QUESTION} }
            
          \end{array}$
       }\\
         \bottomrule
        \multicolumn{4}{m{24cm}}{
          $\begin{array}{rl}
              \textbf{Prompt for AVDN:} & \text{The images first show the drone view and red arrow shows current direction of the drone. How could the drone reach the destination} \\
              & \text{in green bounding box? Note, north is up. Give instruction to the drone on {\$QUESTION}? Concisely describe landmarks but do not} \\
              & \text{mention red arrow and green box.} 
          \end{array}$
       }\\
         \bottomrule
    \end{tabular}}
    \caption{Text prompts for mPLUG-Owl. In RDH task, \$QUESTION is the human inquires from the validation set. In RdI task, \$QUESTION is generated by task performer agent in real-time.}
    \label{tab:appendix_prompt_llm}
\end{table*}

\section{Implementation Details}

\subsection{Helper Model}
\paragraph{SeeRee}
For the training of SeeRee, we apply Mask Language Modeling (MLM) \cite{devlin2018bert} to the response where $80\%$ tokens are masked with [MSK] tokens, and $10\%$ tokens are changed randomly. Cross entropy loss is applied to predictions for the masked tokens:
\begin{equation}
\label{eq_1}
L_{MLM} = \Sigma_{i}L_{Cross Entropy}( y_i, \hat{y}_i ),
\end{equation}
where $y_i$ is the masked token at position $i$ and $\hat{y}_i$ is the prediction.
Additionally, in order to let the COS attention $c$ mask attend to the specific details of the Visual Embedding (VE) that are most relevant to the task, we enforce $C$ to be sparse, letting the VE to sparsely attend to itself using a sparsity loss \cite{lin2022swinbert}:
\begin{equation}
    L_{SPARSE} = \lambda \times \sum_{i=1}^M \sum_{j=1}^M\left|C_{i, j}\right|,
\end{equation}
where $\lambda$ is a regularization hyperparameter and $C_{i,j}$ is the value of the learnable conditional mask $C$.

SeeRee is trained on CVDN \cite{thomason2020vision}, DialFRED \cite{gao2022dialfred} and AVDN \cite{fan2022aerial} datasets individually using AdamW optimizer \cite{loshchilov2018decoupled} for 20k iterations with a batch size of 6 and learning rate of $1e^{-4}$. The training data used are converted from the original training set from each dataset, with the environment-specific scripted sampler that generates image sequences with oracle environment information. We select the trained weights based on the RDH task evaluation result. Training takes about 12 hours on one NVIDIA A6000 GPU.

\paragraph{Multi-modal Large Language Model}
We utilized mPLUG-Owl \cite{ye2023mplugowl} as a representative method for Large Language Model in this paper. mPLUG-Owl takes images and text as input and output natural language. It achieved state-of-the-art performance in various multi-modality tasks. Providing the further path in the form of images and a proper prompt, mPLUG-Owl directly outputs the guidance for the performer agent. Table~\ref{tab:appendix_prompt_llm} shows prompt templates. The QUESTION is relevant to the current task, and it is not strictly formed. For example, it could be ``Should I continue forward?'' in CVDN, ``What does the object look like?'' for DialFRED, ``I am on top of a building block, Can I see the destination?'' for AVDN.

\begin{table*}[t]
    \centering
    \setlength{\abovecaptionskip}{0.1cm}
    \resizebox{\textwidth}{!}{
    \begin{tabular}{llll}
    \toprule
        \multicolumn{4}{m{24cm}}{
          $\begin{array}{rl}
              \textbf{Prompt for CVDN:} & \text{Commander says: `Yes to the kitchen. Go to the left of the fireplace and then all the way up the stairs.' } \\
              & \text{Step by step: 1. Yes. 2. go to the kitchen 3. go the left of the fireplace, 4. go upstairs. Commander says:\_\_\_\_}\\
              &\text{Step by step: 1.}
          \end{array}$
       }\\
           \toprule
        \multicolumn{4}{m{24cm}}{
          $\begin{array}{rl}
              \textbf{Prompt for AVDN:} & \text{Commander says: `Hi drone, if you go four o'clock, and go to a small building, go forward to the bigger building, your destination.' } \\
              & \text{Step by step: 1. Go four o'clock. 2. fly to small building. 3. destination is the bigger building.} \\
              &\text{Commander says: `Hey drone, if fly over the white building.' } \\
              & \text{Step by step: 1. fly over the white building.} \\
              &
              \text{Commander says:\_\_\_\_.} \\
              &\text{Step by step: 1.}
          \end{array}$
       }\\
           \toprule
          \textbf{Dataset}&\multicolumn{1}{c}{\textbf{Original response}} & \multicolumn{2}{c}{\textbf{Processed response}} \\
          \toprule
          
          CVDN &\multicolumn{1}{l}{Yeah keep going around the outside till you get to the end. And sorry about the mixup at first.} & \multicolumn{2}{l}{$\begin{array}{l}
               \text{1. Yeah.}\\
                \text{2. Keep going around the outside.}\\ 
                \text{3. Get to the end.}
          \end{array}$} \\
          \cmidrule(lr){2-4} 
          
          AVDN &\multicolumn{1}{l}{Hey drone, head southwest, go through a small vegetation and arrive at a house that is the final destination.} & \multicolumn{2}{l}{$\begin{array}{l}
               \text{1. head southwest.}\\
               \text{2. go through a small vegetation.} \\
               \text{3. final destination is a house.}
          \end{array}$} \\
         \bottomrule
    \end{tabular}}
    \caption{Examples of our Parse by Step method on the CVDN dataset~\cite{thomason2020vision} and AVDN dataset \cite{fan2022aerial}. We fill the original response to the blank of the prompt and input to GPT-3 for sentence completion. The output from the GPT-3 becomes the processed response with mainly task-related instructions kept and organized in steps. Through Parse by Step, we preprocess the response in the training set.}
    \label{tab:table_00}
\end{table*}
\paragraph{RMM $\mathcal{G}$}
RMM $\mathcal{G}$ is an LSTM-based model that generates natural language response based on the input image sequence and dialog history. We train RMM $\mathcal{G}$ from scratch with the same data used for fine-tuning SeeRee, using a batch size of 8 and a learning rate of $1e^{-4}$. 

\subsection{Task Performer Agents}
We aim to leverage the best available task performers in R2H benchmark. 

For CVDN dataset, we select History Aware Multimodal Transformer (HAMT) \cite{chen2021hamt} as the task performer agent in RDH task. HAMT model is designed for Vision-and-Language Navigation (VLN) tasks. The HAMT model incorporates a long-horizon history into multi-modal decision making. It efficiently encodes all past panoramic observations via a hierarchical vision transformer and then combines text, history, and current observation to predict the next action. The model is first trained end-to-end using several proxy tasks, including single-step action prediction and spatial relation prediction, and then reinforcement learning is used to improve the navigation policy further. As the time we conduct the experiment, HAMT has achieved state-of-the-art results on a broad range of VLN tasks, including CVDN \cite{thomason2020vision}. However, HAMT model is only capable for navigation action prediction and cannot generate questions for asking help. Therefore, we adopt RMM Navigator model ($\mathcal{N}$) with RMM Questioner model ($\mathcal{Q}$) \cite{roman2020rmm} in RdI task, which is the only task performer model to the best of our knowledge that is designed to navigate while communicating in natural language. Both RMM $\mathcal{N}$ and $\mathcal{Q}$ are lstm based models and the RMM Questioner model is designed to generate questions at a fixed interval. It takes into account the Navigator's perspective in the environment and the dialog history.  

For AVDN and DialFRED datasets \cite{fan2022aerial, gao2022dialfred}, we use HAA-Transformer and DialFRED model proposed along with the dataset as the task performer agent because they are still leading the leaderboard \footnote{https://eval.ai/web/challenges/challenge-page/1859/overview} \footnote{https://eval.ai/web/challenges/challenge-page/2049/overview}. DialDRED model and HAA-Transformer model are both based on Episodic Transformer model \cite{pashevich2021episodic}. The DialFRED model trained on DialFRED dataset focuses on three types of questions: location clarification, appearance clarification, and direction clarification. The HAA-Transformer model trained on the AVDN dataset can predict both navigation waypoints and human attention.

\subsection{Prompts and Examples for Parse by Step}
We design different prompts for applying Parse by Step on CVDN dataset \cite{thomason2020vision} and AVDN dataset \cite{fan2022aerial}. Table~\ref{tab:table_00} shows the designed prompt and some example preprocessed results. To process the training set responses, we first insert the original response into the blank of the prompt, which is then inputted into GPT-3 for sentence completion. The output from GPT-3, primarily retaining task-related instructions arranged in steps, is subsequently treated as the processed response. Furthermore, we also eliminate the item numbers to streamline the instructions further. This approach aids in preserving the essential task-related information while enhancing the response's readability and conciseness.

\section{Additional Benchmark Analysis}
We conduct additional analyses on our R2H benchmark across the three datasets as shown in Table~\ref{tab:add_analysis}. Especially, the datasets we have adopted cover a wide range of environment types, including both indoor (CVDN and DialFRED) and outdoor (AVDN) settings, as well as synthetic (DialFRED) and photo-realistic environments (CVDN and AVDN). This diversity adds depth and robustness to our benchmark.

\begin{table*}[t]
    \centering
    \setlength{\abovecaptionskip}{0.1cm}
    \resizebox{\textwidth}{!}{
\begin{tabular}{lcccccc} 
\toprule
& $\begin{array}{l}\text { \# navigation } \\
\text { trajectories }\end{array}$ & $\begin{array}{l}\text { \# querries per } \\
\text { trajectory }\end{array}$ & $\begin{array}{l}\text { Avg length of  } \\
\text { human response (words) }\end{array}$ & $\begin{array}{l}\text { Avg length of  } \\
\text { human query (words) }\end{array}$ & $\begin{array}{l}\text { Avg \# images in input image sequence } \\
\text {of helper agents }\end{array}$ \\
\hline CVDN & 2050 & 2.3 & 14.9 & 9.7 & 11.8 \\
\hline DialFRED & 120958 & 1.0 & 6.0 & 7.0 & 10.8 \\
\hline AVDN & 5372 & 1.8 & 19.7 & 9.0 & 9.1\\
\bottomrule

\end{tabular}
}
\caption{Statistic analysis on R2H benchmark across the three datasets.  The columns from left to right are the number of navigation trajectories; the number of queries per trajectory, which is the same as the number of responses per trajectory; the average length of responses and queries provided by human annotators and the average number of images in the image sequences sampled from the environment by script-based samplers, which serve as input to the helper agent.}
\label{tab:add_analysis}
\end{table*}

\section{RDH Task Examples}
\label{case}
In this section, we show examples of our RDH task on all three datasets, CVDN dataset \cite{thomason2020vision}, DialFRED dataset \cite{gao2022dialfred} and AVDN dataset \cite{fan2022aerial}. As shown in Figure~\ref{fig:case1}, \ref{fig:case2} and \ref{fig:case3} each example includes the input inquiry from human, the sampled input image sequence from the environment oracle information, the response generated by SeeRee and the human ground truth response.

\section{Results of RDH Task on More Metrics}
\label{More_RDH}
Table~\ref{tab:results_RDH} shows the performance of RDH task on the main metric. Here Table~\ref{tab:appendix_table_0} exhibits the results on more metrics for a thorough performance understanding. 

Goal Progress (GP) evaluates the distance of the progress made towards the destination. Success Rate (SR) shows the ratio of tasks being completed successfully. Success weighted by inverse Path Length (SPL) and Path Weighted Success Rate (PWSR) measure the Success Rate weighted by the total length of the navigation trajectory.
Here we could draw similar conclusions with Table~\ref{tab:results_RDH}. Note that AVDN
dataset~\cite{fan2022aerial} is more challenging due to the drone simulator's adoption of a continuous control space and complex visual information. The mistakes made by the helper are likely to be amplified if the task performer misses the target location and overshoots, or if the initial direction provided by helper agent is inaccurate. Even though all methods have a negative impact on the goal progress, SeeRee minimized the influence while achieving the highest Success Rate.

\section{Additional LLM Baselines}
\label{add_LLM_Baseline}
Comprehensive baselines are essential for validating our approach, particularly in light of the growing focus on Large Language Models (LLMs). Given the complexity of our tasks, the multimodal LLM (mPLUG-Owl) serves as an adequate LLM baseline as it inherently supports our task requirements by natively accommodating both image sequences and text for language generation. Additionally, we have also explored two additional baselines that leverage the power of LLMs as detailed below and the result is shown in Table~\ref{tab:add_baselines}.

\begin{table*}[t]
    \centering
    \setlength\tabcolsep{0pt}
    \setlength{\abovecaptionskip}{0.1cm}
    \resizebox{\textwidth}{!}{
\begin{tabular}{lcccccc}
\hline & $\begin{array}{c}\text { CVDN } \\
\text { validation seen } \\
\text{GP}\end{array}$ & $\begin{array}{c}\text { CVDN } \\
\text { validation unseen }\\
\text{GP} \end{array}$ & $\begin{array}{c}\text { DialFRED } \\
\text { validation seen }\\
\text{SR} \end{array}$ & $\begin{array}{c}\text { DialFRED } \\
\text { validation unseen }\\
\text{SR}\end{array}$ & $\begin{array}{c}\text { AVDN } \\
\text { validation seen }\\
\text{SPL}\end{array}$ & $\begin{array}{c}\text { AVDN } \\
\text { validation unseen }\\
\text{SPL}\end{array}$ \\
\hline Multimodal LLM baseline  (mPLUG-Owl)  & 5.3 & \textbf{3.6} & \textbf{47.0} & \textbf{33.8} & 0.7 & 1.5 \\
\hline mPLUG-Owl + ChatGPT & 5.3 & 3.5 & 43.5 & 31.8 & 1.2 & 1.6 \\
\hline BLIP2 & \textbf{5.6} & 3.4 & 38.6 & 24.7 & \textbf{3.8} & \textbf{5.0} \\
\hline
\end{tabular}
}
\caption{Evaluation of multimodal LLM baselines on the RdH task. Additional LLM baselines (bottom two rows) are compared with regard to the multimodal LLM baseline using mPLUG-Owl.}
\label{tab:add_baselines}
\end{table*}

\paragraph{Multimodal-LLM + ChatGPT} We experiment with prompting the Multimodal-LLM (mPLUG-Owl) to generate only captions for the image sequences. Subsequently, we use ChatGPT to serve as the helper agent, generating responses based on the captions and queries from the task performer.

\paragraph{BLIP2 Model with stacked images input} We employ the BLIP2 model \cite{li2023blip} in a zero-shot manner. BLIP2 model benefits from a generic and efficient pre-training strategy that combines pretrained vision models with LLMs for vision-language pretraining. Due to the model's limitation of accepting only a single image input, we stack the input image sequences into a single image arranged in four columns. The text prompt used is the same as the prompt for Multimodal-LLM (mPLUG-Owl) as shown in Table~\ref{tab:appendix_prompt_llm}.

\section{Human Evaluation Details}
\begin{figure*}[t]
    \centering
    \subfloat[Example 1]{\includegraphics[width=0.5\linewidth]{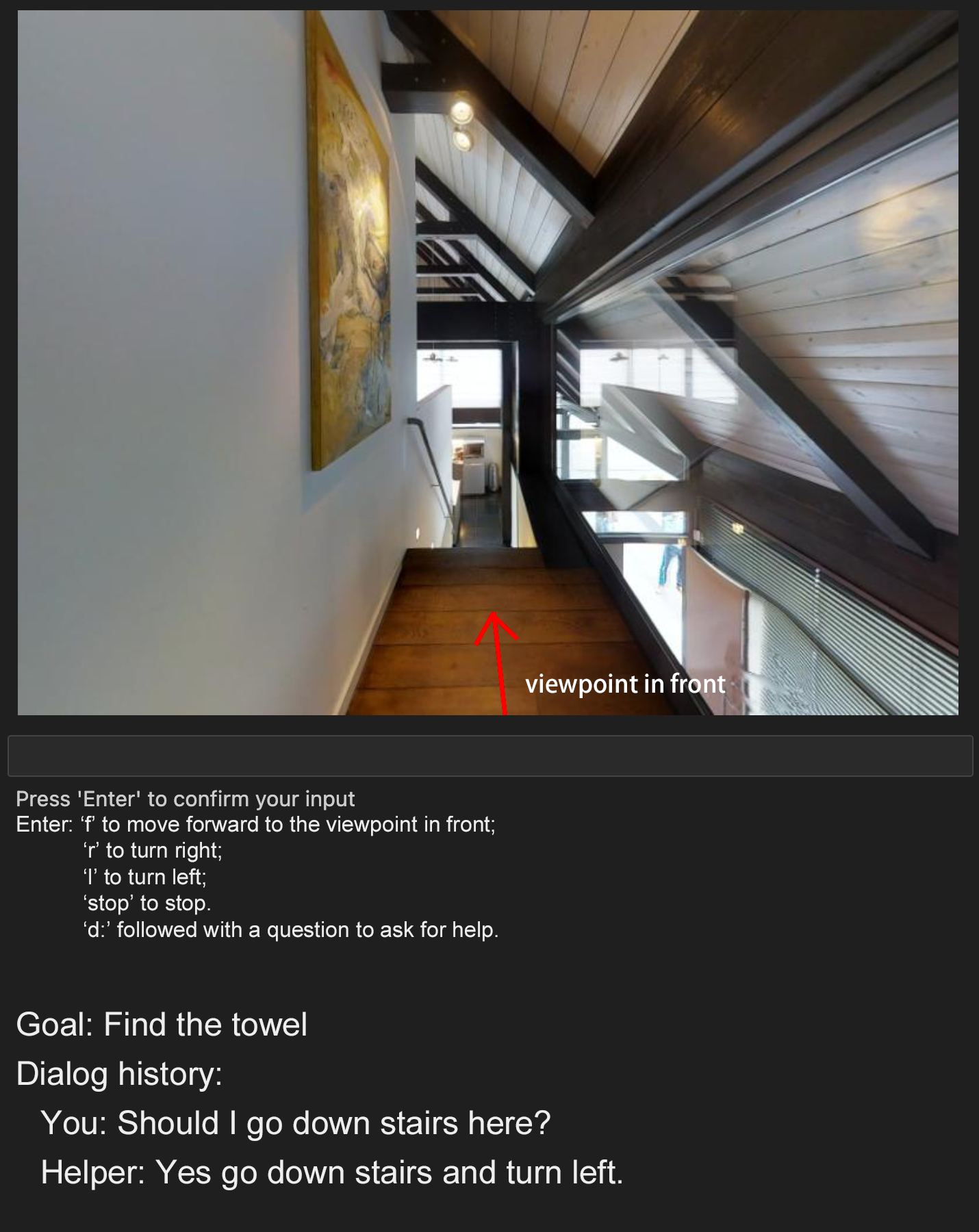}
    }
    \subfloat[Example 2]{\includegraphics[width=0.5\linewidth]{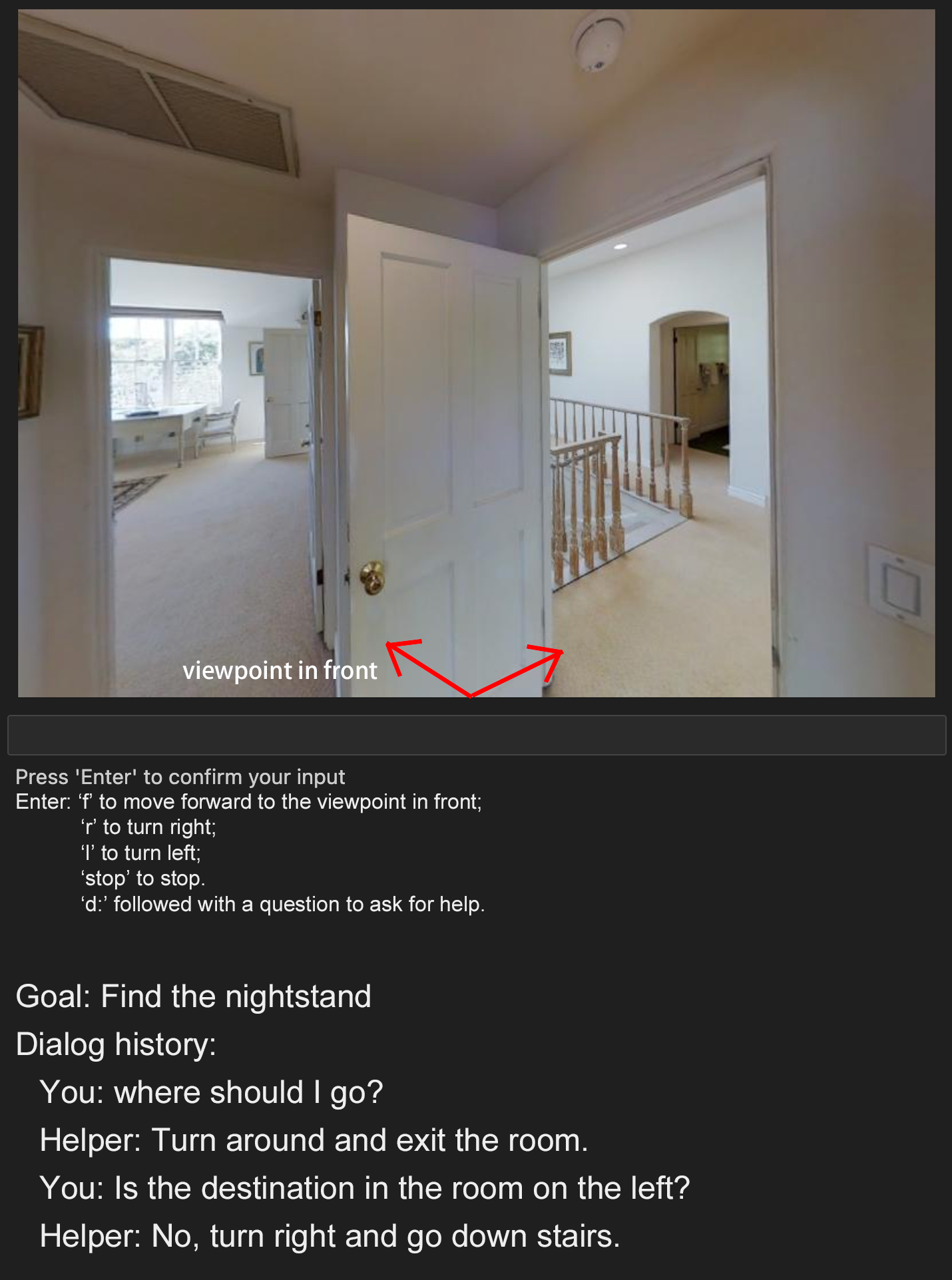}}
    \caption{Examples of the interface used during human evaluation based on RdI task with CVDN \cite{thomason2020vision} dataset.\label{fig:ui}}
\end{figure*}

\begin{figure*}
    \centering

    \includegraphics[width=0.8\linewidth]{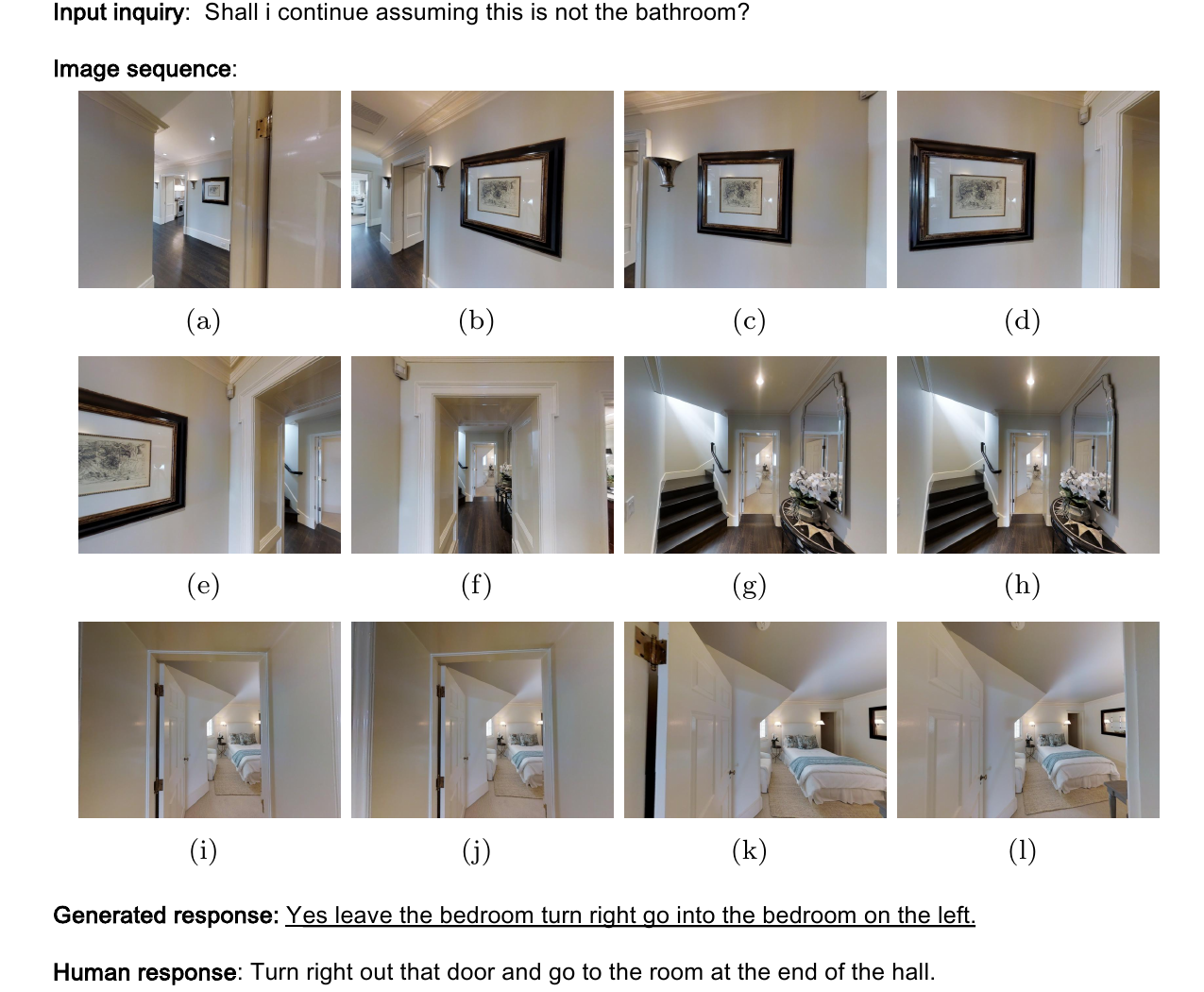}

    \caption{An example of data in RDH task on CVDN dataset. \\Human inquiry: Shall I continue assuming this is not the bathroom? Ground truth human response: Yes leave the bedroom turn right go into the bedroom on the left.\label{fig:case3}} 
\end{figure*}

\begin{figure*}
    \centering

    \includegraphics[width=0.85\linewidth]{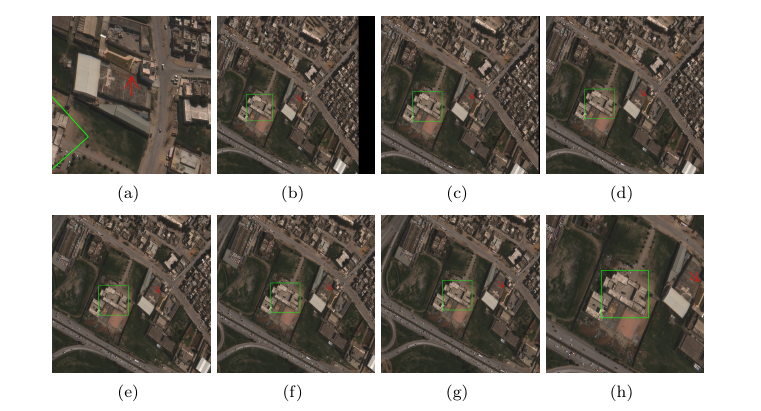}

    \caption{An example of data in RDH task on the AVDN dataset. \\Human inquiry: I see some warehouses in my view. Am I near the destination? How to go to destination? \\Ground truth human response: You are not close to your destination yet, head north and you will reach the complex gray warehouse office.\label{fig:case2}} 
\end{figure*}

\begin{figure*}
    \centering
    \includegraphics[width=0.8\linewidth]{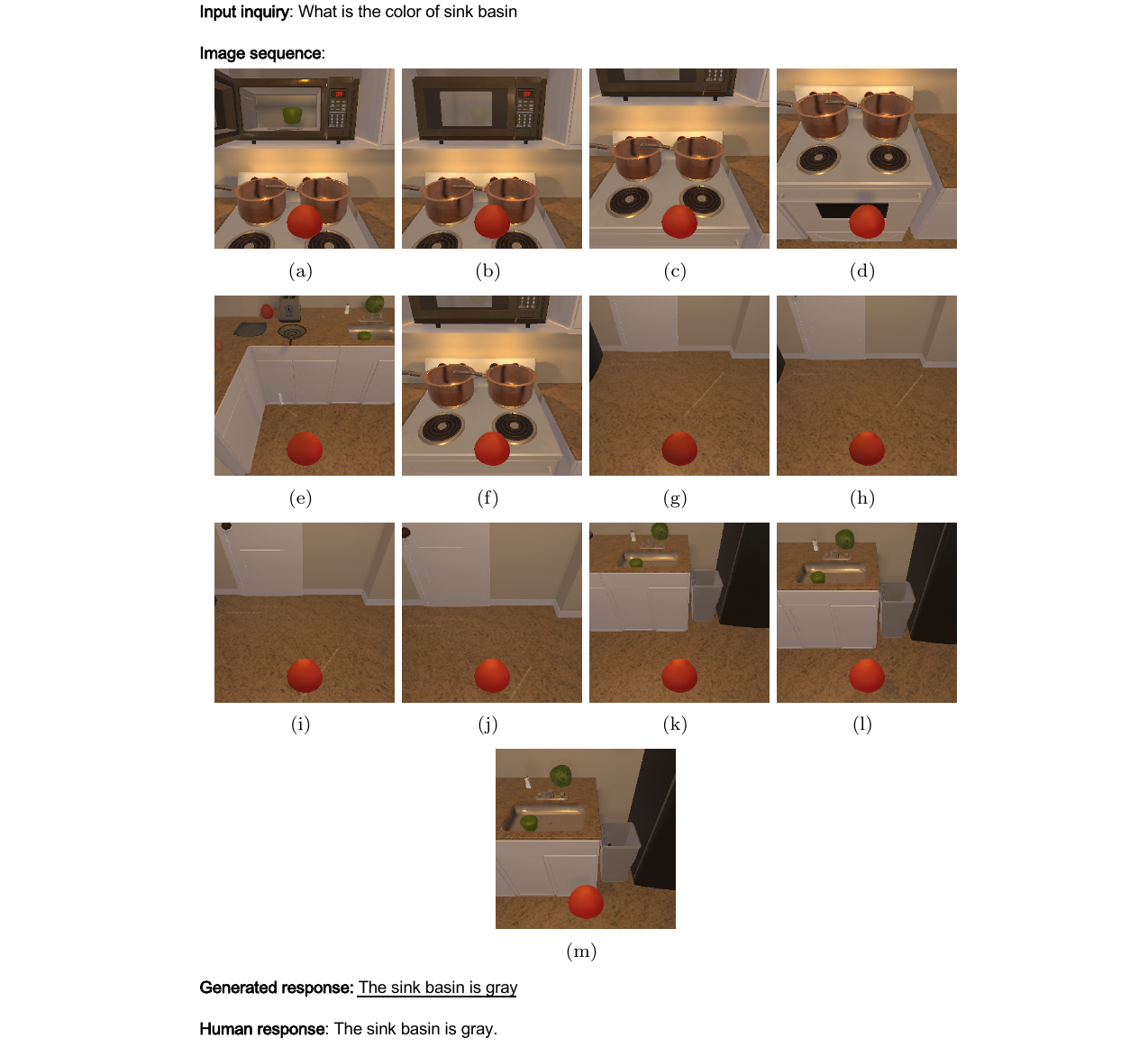}

   \caption{An example of data in RDH task on DialFRED dataset. \\Human inquiry: What is the color of the sink basin? Ground truth human response: The sink basin is Grey.\label{fig:case1}}
\end{figure*}

In our human evaluation study, we recruited five students from the university and paid with at least \$17/h as the human task performer. They consented to participate and contribute their data used for this work. These participants were given instructions on how to use the simulator before the evaluation began and had the same prior knowledge about the project. The participants are randomly assigned a navigation-helper agent and CVDN data including the initial instruction, starting, and target location. An interface, as shown in Figure \ref{fig:ui}, is built where an image box will show the image returned from the simulator and a text box allows both navigation control commands and sending natural language inquires for helper agents. It enables an easy interaction among the human task performer, helper agent, and simulator so that the participants can experience a real-time collaboration between helper and performer with only a screen and keyboard. Upon completion of the task, we assess task success and goal progress in the same way as in \citet{thomason2020vision} and we ask the participant to rate the naturalness and faithfulness of the response from the helper in the last task session.

\begin{table}[t]
\setlength{\abovecaptionskip}{0.1cm}
    \centering

\begin{subtable}[t]{0.48\textwidth}
\resizebox{\columnwidth}{!}{
\setlength\tabcolsep{0pt}
\begin{tabular}{llc|ccc|ccc} 
\toprule
\multicolumn{3}{c|}{}& \multicolumn{3}{c|}{$\begin{array}{c}
     \textbf{Seen}  \\
     \textbf{Validation}
\end{array}$} & \multicolumn{3}{c}{$\begin{array}{c}
     \textbf{Unseen}  \\
     \textbf{Validation}
\end{array}$}\\
 \cmidrule(lr){4-6} 
 \cmidrule(lr){7-9} 

\multicolumn{2}{c|}{\textbf{Helper}} &
\multicolumn{1}{c|}{$\begin{array}{c}
     \textbf{Task}\\
     \textbf{Performer}
\end{array}$}
& \multicolumn{1}{r}{$\begin{array}{r}
     \text{GP}\uparrow
\end{array}$}
& \multicolumn{1}{r}{$\begin{array}{r}
     \text{SPL}\uparrow
\end{array}$}
& \multicolumn{1}{r|}{$\begin{array}{r}
     \text{SR}\uparrow
\end{array}$} & \multicolumn{1}{r}{$\begin{array}{r}
     \text{GP}\uparrow
\end{array}$}& \multicolumn{1}{r}{$\begin{array}{r}
     \text{SPL}\uparrow
\end{array}$} & \multicolumn{1}{r}{$\begin{array}{r}
     \text{SR}\uparrow
\end{array}$}\\
\midrule
\multicolumn{2}{l|}{$\begin{array}{l}
    \text{Human Annotator}\\
\end{array}$} & \multirow{4}{*}{$\begin{array}{l}
    \text{HAMT}
\end{array}$}  & 6.9 & 17.6 & 20.7 & 5.1 & 11.2 & 17.1  \\
\cmidrule(lr){0-1}
\cmidrule(lr){4-9}

 \multicolumn{2}{l|}{$\begin{array}{c}
    \text{RMM }\mathcal{G}
\end{array}$} &  & 4.7 & 8.6 & 14.4 & 2.8 & 6.8 & 10.7  \\

 \multicolumn{2}{l|}{$\begin{array}{c}
    \text{Multimodal LLM}
\end{array}$} &  & 5.3 & \textbf{14.7} & 17.5 & 3.6 & 8.2 & 13.1  \\

 \multicolumn{2}{l|}{$\begin{array}{l}
    \text{SeeRee}\\
\end{array}$} &   & \textbf{6.5} & 14.0 & \textbf{17.8} & \textbf{4.9} & \textbf{10.1} & \textbf{15.2}  \\

\bottomrule
\end{tabular}
}
\caption{CVDN dataset}
\end{subtable}

\begin{subtable}[t]{0.48\textwidth}
\setlength\tabcolsep{2pt}
\resizebox{\columnwidth}{!}{
\begin{tabular}{llc|cc|cc} 
\toprule
\multicolumn{3}{c|}{}& \multicolumn{2}{c|}{$\begin{array}{c}
     \textbf{Seen}  \\
     \textbf{Validation}
\end{array}$} & \multicolumn{2}{c}{$\begin{array}{c}
     \textbf{Unseen}  \\
     \textbf{Validation}
\end{array}$}\\
 \cmidrule(lr){4-5} 
 \cmidrule(lr){6-7} 

\multicolumn{2}{c|}{\textbf{Helper}} &
\multicolumn{1}{c|}{$\begin{array}{c}
     \textbf{Task}\\
     \textbf{Performer}
\end{array}$}
& \multicolumn{1}{r}{$\begin{array}{r}
     \text{SR}\uparrow
\end{array}$}
& \multicolumn{1}{r|}{$\begin{array}{r}
     \text{PWSR}\uparrow
\end{array}$}
& \multicolumn{1}{r}{$\begin{array}{r}
     \text{SR}\uparrow
\end{array}$} & \multicolumn{1}{r}{$\begin{array}{r}
     \text{PWSR}\uparrow
\end{array}$}\\
\midrule
\multicolumn{2}{l|}{$\begin{array}{l}
    \text{Human Annotator}\\
\end{array}$} & \multirow{4}{*}{$\begin{array}{c}
    \text{DialFRED}\\
    \text{Model}
\end{array}$}  & 49.1 & 39.1 & 33.4 & 14.6  \\
\cmidrule(lr){0-1}
\cmidrule(lr){4-7}

 \multicolumn{2}{l|}{$\begin{array}{c}
    \text{RMM }\mathcal{G}
\end{array}$} &  & 46.5 & 35.3 & 32.1 & 13.8  \\
\multicolumn{2}{l|}{$\begin{array}{c}
    \text{Multimodal LLM}
\end{array}$} &  & 47.0  & 32.5 & \textbf{33.8}  & 14.2 \\
 \multicolumn{2}{l|}{$\begin{array}{l}
    \text{SeeRee}\\
\end{array}$} &   & \textbf{49.1} & \textbf{39.1} & 33.1 & \textbf{14.3}  \\

\bottomrule
\end{tabular}
}
\caption{DialFRED dataset}
\end{subtable}

\begin{subtable}[t]{0.48\textwidth}
\resizebox{\columnwidth}{!}{
\setlength\tabcolsep{0pt}
\begin{tabular}{llc|ccc|ccc} 
\toprule
\multicolumn{3}{c|}{}& \multicolumn{3}{c|}{$\begin{array}{c}
     \textbf{Seen}  \\
     \textbf{Validation}
\end{array}$} & \multicolumn{3}{c}{$\begin{array}{c}
     \textbf{Unseen}  \\
     \textbf{Validation}
\end{array}$}\\
 \cmidrule(lr){4-6} 
 \cmidrule(lr){7-9} 

\multicolumn{2}{c|}{\textbf{Helper}} &
\multicolumn{1}{c|}{$\begin{array}{c}
     \textbf{Task}\\
     \textbf{Performer}
\end{array}$}
& \multicolumn{1}{r}{$\begin{array}{r}
     \text{GP}\uparrow
\end{array}$}
& \multicolumn{1}{r}{$\begin{array}{r}
     \text{SPL}\uparrow
\end{array}$}
& \multicolumn{1}{r|}{$\begin{array}{r}
     \text{SR}\uparrow
\end{array}$} & \multicolumn{1}{r}{$\begin{array}{r}
     \text{GP}\uparrow
\end{array}$}& \multicolumn{1}{r}{$\begin{array}{r}
     \text{SPL}\uparrow
\end{array}$} & \multicolumn{1}{r}{$\begin{array}{r}
     \text{SR}\uparrow
\end{array}$}\\
\midrule
\multicolumn{2}{l|}{$\begin{array}{l}
    \text{Human Annotator}\\
\end{array}$} & \multirow{4}{*}{$\begin{array}{c}
    \text{HAA-}\\
    \text{Transformer}
\end{array}$}  & 56.3 & 14.7 & 17.3 & 55.2 & 16.5 & 20.4  \\
\cmidrule(lr){0-1}
\cmidrule(lr){4-9}

 \multicolumn{2}{l|}{$\begin{array}{c}
    \text{RMM }\mathcal{G}
\end{array}$} &  & -37.0 & 2.4 & 3.5 & -26.6 & 3.6 & 4.9  \\

 \multicolumn{2}{l|}{$\begin{array}{c}
    \text{Multimodal LLM}
\end{array}$} &  & -96.5 & 0.7 & 0.8 & -95.9 & 1.5 & 2.2  \\

 \multicolumn{2}{l|}{$\begin{array}{l}
    \text{SeeRee}\\
\end{array}$} &   & \textbf{-29.6} & \textbf{4.6} & \textbf{5.7} & \textbf{-24.9} & \textbf{4.4} & \textbf{6.1}  \\

\bottomrule
\end{tabular}
}
\caption{AVDN dataset}
\end{subtable}

\caption{Results of Respond to Dialog History (RDH) task on CVDN dataset (a), DialFRED dataset (b) and AVDN dataset \cite{thomason2020vision,gao2022dialfred, fan2022aerial}. We replace the response in validation sets with responses from different helper agents. Since the task performer agent is trained with human dialog and is fixed, it relies on the information contained in the response to complete the task. Thus, the better the task performer agent performs in the task, the more effective the response is.}
\label{tab:appendix_table_0}
\end{table}

\end{document}